\documentclass[11pt]{article}
\usepackage{rotating}
\usepackage{acl2015}
\usepackage{times}
\usepackage{url}
\usepackage{latexsym}
\usepackage{tabularx}
\usepackage{makecell}
\usepackage{hyperref}
\usepackage{multirow}
\usepackage{multicol}
\usepackage{booktabs}
\usepackage{CJK}
\usepackage[section]{placeins}
\usepackage[hang,flushmargin]{footmisc} 
\usepackage[T1]{fontenc}
\usepackage{lipsum}
\newcommand\blfootnote[1]{%
  \begingroup
  \renewcommand\thefootnote{}\footnote{#1}%
  \addtocounter{footnote}{-1}%
  \endgroup
}
\usepackage{fancyhdr}
\title{A Comparative Study between Full-Parameter and LoRA-based Fine-Tuning on Chinese Instruction Data for Instruction Following Large Language Model}

\author{Xianghui Sun, Yunjie Ji, Baochang Ma{*}, Xiangang Li \\
Beike Inc., Beijing, China  \\
\texttt{\{sunxianghui002,jiyunjie001,mabaochang001,lixiangang002\}@ke.com}}

\begin{document}
\maketitle
\begin{abstract}
Recently, the instruction-tuning of large language models is a crucial area of research in the field of natural language processing. Due to resource and cost limitations, several researchers have employed parameter-efficient tuning techniques, such as LoRA, for instruction tuning, and have obtained encouraging results. In comparison to full-parameter fine-tuning, LoRA-based tuning demonstrates salient benefits in terms of training costs. In this study, we undertook experimental comparisons between full-parameter fine-tuning and LoRA-based tuning methods, utilizing LLaMA as the base model.

The experimental results show that the selection of the foundational model, training dataset scale, learnable parameter quantity, and model training cost are all important factors. We hope that the experimental conclusions of this paper can provide inspiration for training large language models, especially in the field of Chinese, and help researchers find a better trade-off strategy between training cost and model performance.
To facilitate the reproduction of the paper's results, the dataset, model and code will be released.\textsuperscript{1}\blfootnote{
    \textsuperscript{*}Corresponding author \\
\textsuperscript{1}https://github.com/LianjiaTech/BELLE
}.
\end{abstract}

\section{Introduction}

The advent of language models such as ChatGPT\cite{Chatgpt} and GPT-4\cite{Gpt-4}, which exhibit human-like understanding and generation capabilities across various domains, has highlighted the importance of instruction tuning in enabling these models to better comprehend human instructions. 
Currently, there exist several open-source, large language models that have been fine-tuned on instructional data, including OPT\cite{OPTOpenPretrained2022}, BLOOM\cite{BLOOM176BParameterOpenAccess2022}, LLaMA\cite{touvron2023llama}, and GLM\cite{zeng2023glm-130b}. These models have demonstrated exceptional performance on a range of language tasks, thereby underscoring the potential benefits of instruction tuning in enhancing language model performance.

In the field of model training, two widely used methods are full-parameter fine-tuning and parameter-efficient tuning. Recently, researchers have conducted extensive experiments to compare the effectiveness of various parameter-efficient tuning methods such as Adapters \cite{AdapterFirst2019,adapater2}, LoRA \cite{LoRA}, and P-tuning \cite{PrefixTuning2021,pTuning,ptuning2} against full-parameter fine-tuning \cite{Liupeftoverview}. The results of these experiments demonstrate that LoRA is a promising parameter-efficient tuning method and has been applied in many studies to fine-tune large language models with significant success \cite{alpaca-lora,Baize}.

However, the effectiveness and efficiency of LoRA for finetuning a instruction-following model have not been well explored. In this paper, we examined the influence of two factors: base model and training data scale. Besides, we also compared LoRA with full-parameter finetuning from the perspective of model performance and training efficiency. We assessed these models on a evaluation set consisting of 1,000 samples, spanning across 9 real-word use cases. Finally we obtained the following important experimental results:

\begin{itemize}
\item 
The choice of the base model has a significant impact on the effectiveness of LoRA-based tuning. 
\item
Increasing the amount of training data can continuously improve the model’s effectiveness
\item
LoRA-based tuning benefits from the number of model parameters
\end{itemize}

We hope that the experimental conclusions of this paper can provide inspiration for training large language models, especially in the field of Chinese, and help researchers find a better trade-off strategy between training cost and model performance.

\section{Related work}

\subsection{Instruction tuning}
Recent studies\cite{PaLMScalingLanguage2022,OPTOpenPretrained2022} have found that by fine-tuning models on datasets with human-annotated prompts, known as instruction-tuning, models can execute new tasks by understanding task instructions, thereby improving their zero-shot and few-shot generalization abilities on unseen tasks. Early research focused on instruction tuning a general NLP task solver, and there is a trend towards converting more and more NLP datasets into a unified dataset and then conducting multi-task training \cite{xu2022zeroprompt,xie2022unifiedskg,wang2022super,khashabi2020unifiedqa,min2021metaicl,ye2021crossfit,liu2019multi,zhong2021adapting,ScalingInstructionFinetunedLanguage2022}. Some research efforts even employ reinforcement learning from human feedback (RLHF) strategies to make models more adherent to human instructions.\cite{TrainingLanguageModels2022,ConstitutionalAIHarmlessness2022,FineTuningLanguageModels2020,LearningSummarizeHuman2022,WebGPTBrowserassistedQuestionanswering2022,PretrainingLanguageModels2023} 
Today, instruction tuning has had a profound impact on the field of natural language processing (NLP). The emergence of technologies such as ChatGPT\cite{Chatgpt} and GPT-4\cite{Gpt-4} has attracted more researchers to engage in the development of instruction tuning. Compared to English instruction data, there is currently less research on instruction tuning on Chinese instruction data, which to some extent hinders the development of large language models in the Chinese field.

\subsection{Parameter-efficient tuning}
As the model size continues to increase, fine-tuning all parameters becomes more challenging since it is necessary to save the gradients and optimizer states for all parameters. 
Therefore, researchers have proposed parameter-efficient tuning, a low-resource and efficient tuning method that only tunes a small number of parameters or introduces additional trainable parameters. Prefix Tuning \cite{pTuning,PrefixTuning2021,ptuning2} add trainable virtual token embeddings and fix the whole model.
Adapters\cite{AdapterFirst2019,adapater2} inserting adapter layers between existing layers in neural networks and only fine-tuning the adapter network's parameters. 

\cite{Aghajanyan} show that the learned over-parametrized models in fact reside on a low intrinsic dimension. \cite{LoRA} Inspired by this work and proposed LoRA approach, which suggests that weights update during model adaptation for downstream tasks should also have a low "intrinsic rank". Experimental results from \cite{Liupeftoverview} suggest that LoRA is a relatively effective method among various parameter-efficient tuning approaches. It has been adopted by many recent open-source projects\cite{alpaca-lora,Baize} for training large language models and achieved promising results. These research works only consider LoRA as a method of training models and does not have an in-depth analysis of factors affecting LoRA-based tuning results.

\section{Method}
In this section, we will provide a brief introduction to LoRA(Low-Rank Adaption)\cite{LoRA}.

For a pre-trained weight matrix $W_0\in R^{d\times k}$, its updates can be represented by a low-rank decomposition:
\begin{equation}
    W_0+\Delta W = W_0 + BA
\end{equation}
where $B\in R^{d\times r}$, $A\in R^{r\times k}$, and the rank $r\ll \min(d,k)$. 
For a linear layer $h = W_0x$, the forward pass is modified to be to be:
\begin{equation}
    h = W_0x+\Delta Wx = W_0x+BAx
\end{equation}

Matrix A will be initialized by random Gaussian and B will be initialized by zero, making the initial value of $\Delta W = BA$ zero at the start of the training. \cite{LoRA} only adapted the attention weights for downstream tasks and freeze the MLP modules, we follow Baize\cite{Baize} which applies LoRA to adapt all linear layers at the same time.

\section{Experiments}
We adopted the datasets constructed in our previous work\cite{belle-data}, selecting three data scales of 0.6M, 2M and 4M respectively. Combining these three datasets, we aim to investigate the impact of different training data sizes on the performance of LoRA-based tuning. To verify whether conducting LoRA-based tuning on the model after instruction tuning can further improve the model performance, we also selected the math\_0.25M dataset, which is a dataset focusing on the mathematical problem-solving field. 

\begin{figure*} [t!]
	\centering
	\includegraphics[scale=0.5]{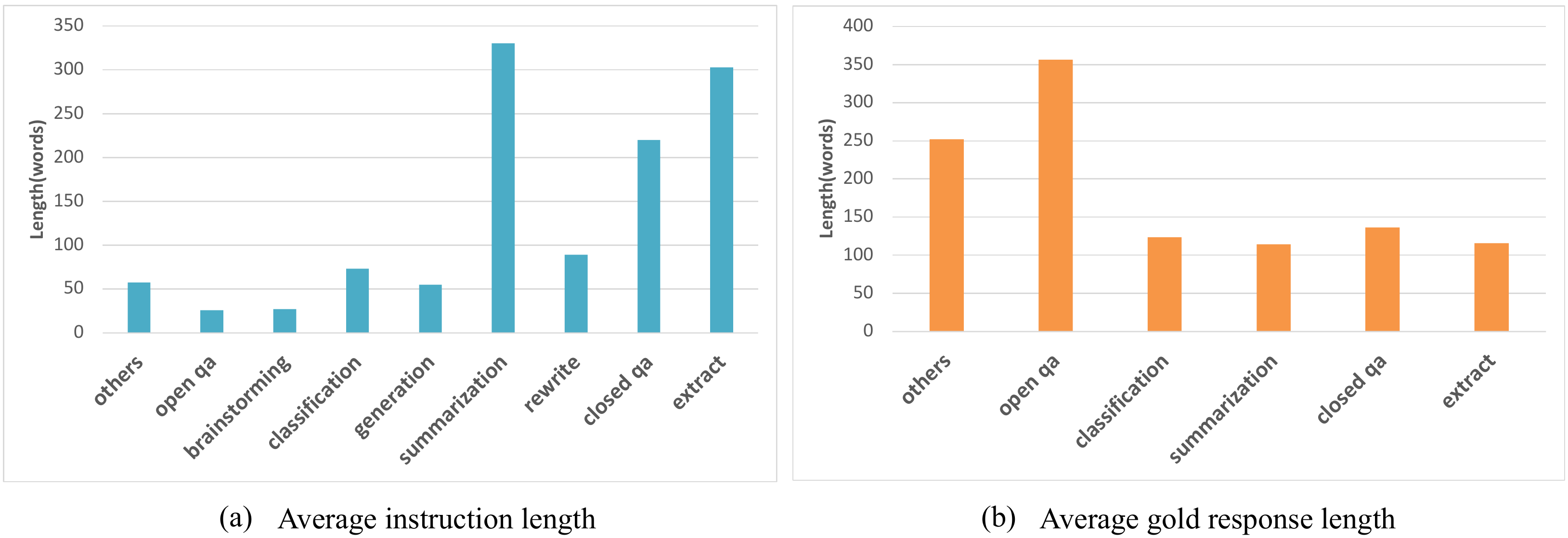}
	\caption{(a) shows average length of instructions, (b) show average length of gold responses.}
\label{length}
\end{figure*}

The evaluate set consists of 1,000 rigorously manually screened and processed data entries, covering nine categories, including translation, Open QA, closed QA, generation, and other tasks closely related to practical applications. Table \ref{data_stat} demonstrates the number of samples in each category of the evaluate set and Figure \ref{length}
shows the length of evaluation samples. The category Other contains two types of data: math and code, where math refers to solving mathematical application problems and code refers to code generation

\begin{table}[t!]
\caption{The number and average prompt length of each type of instructions}
\small
\begin{center}
\begin{tabular}{c|c} 
\hline 
\textbf{Use case} & \textbf{\#Nums}  \\
\hline   
Others  & 113 \\
\hline
Open QA  & 285  \\
\hline
Brainstorming  & 179 \\
\hline
Classification  & 65 \\
\hline
Generation  & 98\\
\hline
Summarization  & 40  \\
\hline
Rewrite  & 131 \\
\hline
Closed QA  & 52 \\
\hline
Extract  & 37 \\
\hline
\end{tabular}
\end{center}
\label{data_stat}
\small
\end{table}

\subsection{Model Settings}
In this study, we selected LLaMA\cite{touvron2023llama} as our foundational experimental models. LLaMA, released by Meta AI, is a collection of large-scale language models with four different parameter scales: 7B, 13B, 33B, and 65B. The performance of LLaMA model is outstanding, with empirical evidence showing that LLaMA-13B, with only 1/10 of the parameter scale, outperforms GPT-3 (175B)\cite{LanguageModelsAre2020} in most benchmark evaluations. In this paper, we chose LLaMA-7B and LLaMA-13B as our base experimental models. 

\begin{table}[t!]
\caption{Hyper-parameter settings of full-parameters fine-tuning}
\begin{center}
\begin{tabular}{l|r} 
\hline 
\textbf{Hyper parameter} & \textbf{Value} \\
\hline   
Precision  & bf16 \\
\hline
Epochs  & 3 \\
\hline
Batch size  & 32 \\
\hline
Learning rate  & 5e-6 \\
\hline
Warmup ratio  & 0.03 \\
\hline
LR scheduler type  & cosine \\
\hline
Max length  & 1024 \\
\hline
\end{tabular}
\end{center}
\label{hyper-parameters-ft}
\end{table}

For the full-parameters fine-tuning experiment, Table \ref{hyper-parameters-ft} list the hyper-parameters of fine-tuning.

For the LoRA experiment, we followed the hyper-parameters in \cite{Baize}, which set the rank in LoRA to 8 and apply LoRA to adapt attention weights and all linear layers, more details in list in Table \ref{hyper-parameters-lora}. This experiment was conducted on 8 NVIDIA A100-40GB GPUs.

\begin{table}[t!]
\caption{Hyper-parameter settings of LoRA-based tuning}
\begin{center}
\begin{tabular}{l|r} 
\hline 
\textbf{Hyper parameter} & \textbf{Value} \\
\hline   
Precision  & fp16 \\
\hline
Epochs  & 4 \\
\hline
Batch size  & 128 \\
\hline
Learning rate  & 2e-4 \\
\hline
Warmup steps  & 100 \\
\hline
LR scheduler type  & cosine \\
\hline
Max length  & 1024 \\
\hline
\end{tabular}
\end{center}
\label{hyper-parameters-lora}
\end{table}

\begin{table*}[t!]
\caption{Main results. In this table, LLaMA-13B + LoRA(2M) represents a model trained on 2M instruction data using LLaMA-13B as base model and LoRA training method, and LLaMA-7B + FT(2M) represents a model trained using full-parameters fine-tuning.
LLaMA-7B + FT(2M) + LoRA(math\_0.25M) represents a model trained on 0.25M mathematical instruction data using LLaMA-7B + FT(2M) as the base model and LoRA training method, and LLaMA-7B + FT(2M) + FT(math\_0.25M) represents a model trained using incremental full-parameters fine-tuning.
About the training time, all these experiments were conducted on 8 NVIDIA A100-40GB GPUs.}
\small
\begin{center}
\begin{tabular}{lccc} 
\hline 
\textbf{Model} & \textbf{Average Score} & \textbf{Additional Param.} & \textbf{Training Time (Hour/epoch)}\\
\hline

LLaMA-13B + LoRA(2M) &0.648 &28M &10\\
LLaMA-7B + LoRA(4M) &0.624 &17.9M &14\\
LLaMA-7B + LoRA(2M) &0.609 &17.9M &7\\
LLaMA-7B + LoRA(0.6M) &0.589 &17.9M &5\\
\hline

LLaMA-7B + FT(2M) &0.710 &- &31 \\
LLaMA-7B + FT(0.6M) &0.686 &- &17 \\
\hline

LLaMA-7B + FT(2M) + LoRA(math\_0.25M) &0.729 &17.9M &2\\
LLaMA-7B + FT(2M) + FT(math\_0.25M) &0.738 &- &4 \\
\hline
\end{tabular}
\end{center}
\label{traing_conf_lora}
\end{table*}

\subsection{Metrics}
ChatGPT is asked to evaluate responses generated by instruction-following models. For all instructions, ChatGPT gives a score between 0 and 1, where score 0 is the worst and score 1 is the best. In order to reduce randomness, we set the temperature to 0.001 for model generation. Evaluation is achieved by invoking gpt-3.5-turbo API at the time of April 15, 2023.
We calculate model’s scores for each task category and derive its overall performance on the evaluation set using macro average across these categories.

Given limitations of ChatGPT in evaluating mathematical and coding tasks, we compute the scores that include all categories (denoted as average\_score). The detailed scores on each task category can be found in the Appendix.

\subsection{Comparison of Base Models and Dataset Scale for LoRA Tuning}

Firstly, we designed an experiment to compare the performance of LoRA-based instruct tuning on instruction datasets of different sizes. We selected datasets of 0.6M, 2M, and 4M, and the experimental results are presented in Table \ref{traing_conf_lora}. 
As can be seen from the results, similar to most learning tasks, as the dataset size increases, the LoRA-based instruct tuned model exhibits better performance in instruction comprehension.

In addition, we also compared the impact of different base models (LLaMA-7B and LLaMA-13B) on performance. It can be seen that the base model with a larger number of parameters brings a significant improvement in performance. Using LLaMA-7B+LoRA(2M) as the base, changing from 7B to 13B resulted in a larger improvement in performance compared to going from 2M to 4M.

In terms of training time, it can also be observed that LLaMA-13B+LoRA(2M) has certain advantages over LLaMA-7B+LoRA(4M). Better training results were achieved with less training time. However, it should be noted that when using these two models for inference, the LLaMA-7B-based model has an advantage in terms of inference speed and cost due to its lower number of global parameters.

\subsection{Comparison between Full-Parameter and LoRA-based Fine-Tuning}

How does the performance of LoRA-based models compare to full-parameters finetuning? 
As a comparison, we trained two models using full-parameters fine-tuning on instruction training data of 0.6M and 2M, and the results are shown in Table \ref{traing_conf_lora}, which are shown as LLaMA-7B + FT(0.6M) and LLaMA-7B + FT(2M).
It can be seen that full-parameters fine-tuning brings better experimental results.

One intuitive understanding or analysis is that the pre-training large language model, which is trained to generate next word, requires a more complex learning task to switch to instruct following.
LoRA's learning method can only change a relatively small number of parameters, which is more challenging compared to changing all parameters.

Sure, there is no free lunch in the world. Compared to LoRA fine-tuning, using full-parameters fine-tuning requires about 3-5 times the time cost to complete the training.

\subsection{Performing LoRA Tuning for Specified Task}

According to our evaluation, details in the appendix, our models did not perform well on math tasks, with scores mostly below 0.5. 
To verify the adaptation capability of LoRA on specific tasks, we used incremental 0.25M math dataset (math\_0.25M) to adapt the instruction-following large language model (We chose LLaMA-7B + FT(2M) as the base model).

As a comparison, we used incremental fine-tuning with a learning rate of 5e-7 and trained for 2 epochs. So we got two models, one is the LLaMA-7B + FT(2M) + LoRA(math\_0.25M), and the other is LLaMA-7B + FT(2M) + FT(math\_0.25M). 

From the experimental results, it can be seen that incremental fine-tuning still showed better performance but took longer training time. 
Both LoRA and incremental fine-tuning improved the overall performance of the model. 
From the detailed data in the appendix, both LoRA and incremental fine-tuning showed significant improvements in the math task while only causing slight decreases in performance in other tasks. 
Specifically, the math task performance improved to 0.586 and 0.559 respectively.

\subsection{Discussion and Conclusions}

In this article, we conducted an experimental comparison between full-parameter fine-tuning and LoRA-based tuning methods using LLaMA as the base model. We also explored the impact of different amounts of training data and model parameters on the effectiveness of LoRA-based tuning. From the experimental results comparison, some interesting ideas can observed:

1) The choice of the base model has a significant impact on the effectiveness of LoRA-based tuning. Comparing LLaMA-7B+LoRA(0.6M) and LLaMA-7B+FT(0.6M), as well as LLaMA-7B+LoRA(2M) and LLaMA-7B+FT(2M), it is evident that LoRA-based tuning on a base model that has not undergone instruction tuning has limited effectiveness and is far less effective than full-parameter fine-tuning (averaging 10 points lower). However, by comparing LLaMA-7B+FT(2M)+FT(math\_0.25M) and LLaMA-7B+FT(2M)+LoRA(math\_0.25M), it can be seen that LoRA-based tuning on a model that has undergone instruction tuning can achieve comparable results to fine-tuning. This indicates that the choice of the base model is crucial to the effectiveness of the LoRA-based tuning method.

2) Increasing the amount of training data can continuously improve the model's effectiveness. Comparing LLaMA-7B+LoRA(0.6M), LLaMA-7B+LoRA(2M), and LLaMA-7B+LoRA(4M) shows that as the amount of training data increases, the model's effectiveness improves (an average of approximately 2 points improvement for every doubling of data).

3) LoRA-based tuning benefits from the number of model parameters. Comparing LLaMA-7B+LoRA(4M) and LLaMA-13B+LoRA(2M) shows that the number of model parameters has a greater impact on the effectiveness of LoRA-based tuning than the amount of training data.


\bibliographystyle{acl}
\bibliography{acl2015}

\section{Appendix A}
\subsection{Detailed evaluation scores}
\label{detailed_scores}
\begin{sidewaystable*} 
  \centering 
  \caption{Detailed scores on each task category. }
  \label{tab:example} 
 \begin{tabular}{cccccccccccc}
 \hline
         \textbf{\makecell[c]{Model}} & \textbf{\makecell[c]{Training\\ data}} & \textbf{others} & \textbf{rewrite} & \textbf{\makecell[c]{classif-\\ication}} & \textbf{generation} & \textbf{\makecell[c]{summari-\\zation}} & \textbf{extract} & \textbf{\makecell[c]{open\\qa}}& \textbf{\makecell[c]{brain-\\storming}} & \textbf{\makecell[c]{closed\\qa}}& \textbf{\makecell[c]{macro\\ave}} \\ 
        \hline
          \makecell[c]{LLaMA-7B+ LoRA} & \makecell[c]{0.6M} & 0.358&0.719&0.695&0.816&0.65&0.448&0.315&0.793&0.51&0.589 \\ 
          \makecell[c]{LLaMA-7B+ LoRA} & \makecell[c]{2M} &0.364&0.795&0.676&0.854&0.617&0.472&0.369&0.808&0.531&0.61 \\
          \makecell[c]{LLaMA-7B+ LoRA} & \makecell[c]{4M} &0.341&0.821&0.677&0.847&0.645&0.467&0.374&0.806&0.639&0.624  \\
          \makecell[c]{LLaMA-13B+ LoRA} & \makecell[c]{2M} &0.422&0.810&0.696&0.837&0.700&0.537&0.435&0.823&0.577&0.648 \\
        \hline
         \makecell[c]{LLaMA-7B+ FT} & 0.6M& 0.438&0.869&0.698&0.917&0.701&0.592&0.477&0.870&0.606&0.686  \\ 
         \makecell[c]{LLaMA-7B+ FT} & 2M& 0.399&0.871&0.775&0.920&0.734&0.603&0.555&0.900&0.633&0.710  \\ 
         \hline
          \makecell[c]{LLaMA-7B + FT(2M)\\ + LoRA} & math0.25M& 0.560&0.863&0.758&0.915&0.754&0.651&0.518&0.886&0.656&0.729    \\ 
         \makecell[c]{LLaMA-7B + FT(2M)\\ + FT}&  math0.25M&0.586&0.887&0.763&0.955&0.749&0.658&0.523&0.872&0.652&0.738   \\
         \hline
\end{tabular}
\end{sidewaystable*}

\end{document}